\documentclass[letterpaper, 10 pt, conference]{ieeeconf}

\IEEEoverridecommandlockouts
\usepackage{cite}
\usepackage{amsmath,amssymb,amsfonts}
\usepackage{graphicx}
\usepackage{algorithm}
\usepackage{algpseudocode}
\usepackage{xcolor}

\definecolor{projectblue}{RGB}{0,82,155}

\usepackage{colortbl}
\usepackage{caption}
\usepackage{bm}
\usepackage{gensymb}
\usepackage{booktabs}
\usepackage{ragged2e}
\usepackage{hyperref}
\usepackage{cleveref}

\hypersetup{
    hidelinks
}

\usepackage{tabularx}
\usepackage{dblfloatfix}
\usepackage{placeins}
\usepackage{url}
\usepackage{etoolbox}
\usepackage{multirow}
\usepackage{multicol}

\DeclareMathOperator{\Exp}{Exp}

\makeatletter
\g@addto@macro\@maketitle{%
  \par\vspace{0.8em}

  \captionsetup{type=figure}
  \setcounter{figure}{0}
  \centering
  \includegraphics[width=2\columnwidth]{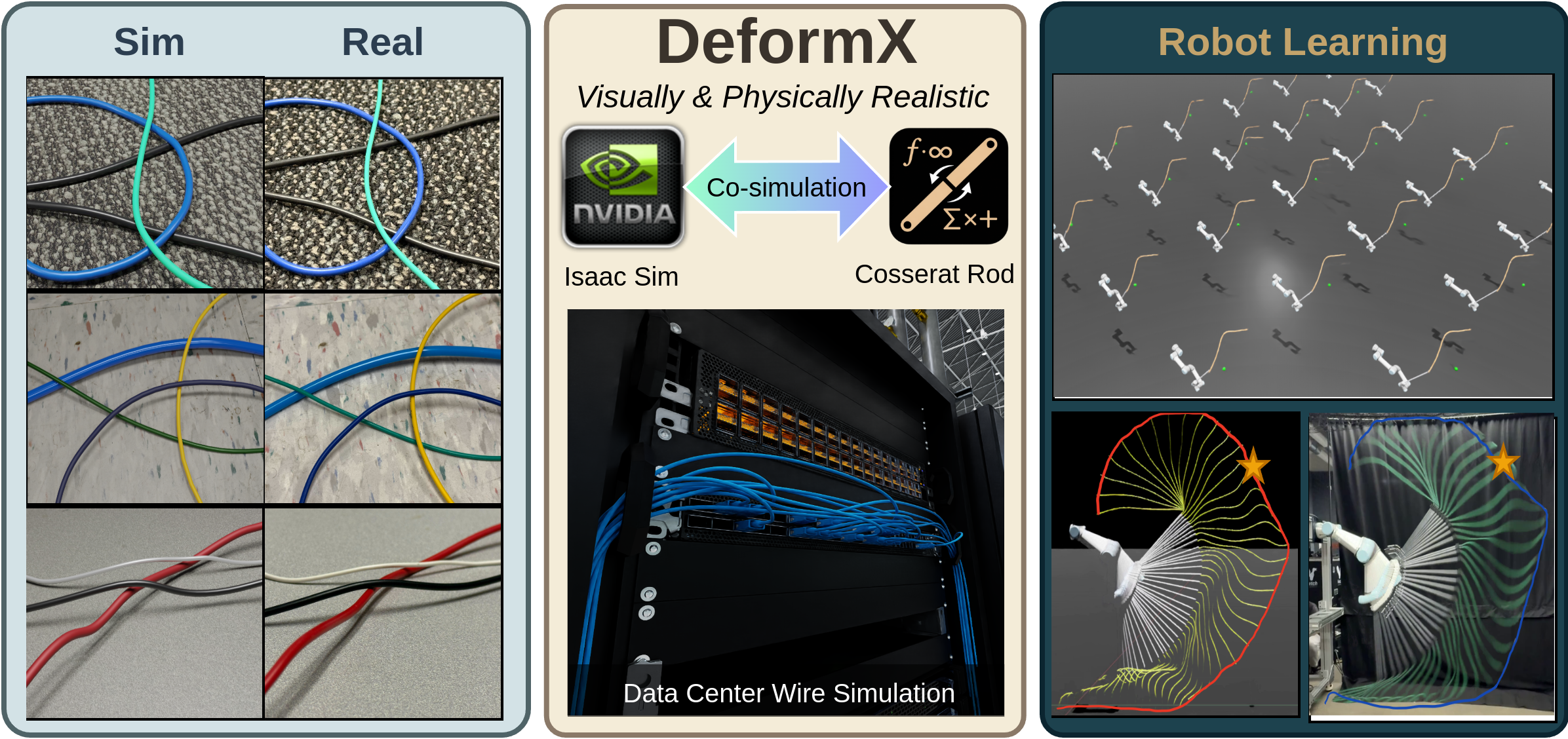}
  \captionof{figure}{
    DeformX is a co-simulation framework for deformable linear objects
    (DLOs) that integrates NVIDIA Isaac Sim with a dedicated Cosserat rod
    engine, providing visually and physically realistic simulation.
    The framework enables scalable data generation and robot learning,
    supporting DLO perception and manipulation.
  }
  \label{fig:figure1}
  \vspace{-5pt}
}
\makeatother

\algrenewcommand\algorithmicindent{1.0em}

\title{
\Large\bfseries
DeformX: A Versatile Co-Simulation Framework for Deformable Linear Objects
\\[-0.15em]
{\small\normalfont\sffamily\bfseries
\href{https://deformx.github.io/}{
  \textcolor{projectblue}{Project Page: deformx.github.io}
}}
\vspace{-1.2em}
}
\author{
Yi Yang$^{1,3,4,\dagger}$,
Xiang Fei$^{1,\dagger}$,
Lehong Wang$^{1,\dagger}$,
Chenhao Li$^{2}$,
Zilin Dai$^{5}$,
Henry Kou$^{1}$,
\\
Lu Li$^{1}$,
Howie Choset$^{1}$
\thanks{$^{\dagger}$ Equal contribution; each reserves the right to be listed first.}
\thanks{$^{1}$ The Robotics Institute, Carnegie Mellon University,
5000 Forbes Ave, Pittsburgh, PA 15213 USA.}
\thanks{$^{2}$ Department of Mechanical Engineering,
Carnegie Mellon University,
5000 Forbes Ave, Pittsburgh, PA 15213 USA.}
\thanks{$^{3}$ School of Ocean and Civil Engineering,
Shanghai Jiao Tong University,
800 Dongchuan Rd, Minhang District, Shanghai, China.}
\thanks{$^{4}$ Zhiyuan College,
Shanghai Jiao Tong University,
800 Dongchuan Rd, Minhang District, Shanghai, China.}
\thanks{$^{5}$ John A. Paulson School of Engineering and Applied Sciences,
Harvard University,
29 Oxford Street, Cambridge, MA 02138 USA.}
}

\begin{document}

\maketitle

\thispagestyle{empty}
\pagestyle{empty}

\begin{abstract}

Deformable linear objects (DLOs) such as wires, cables, and ropes are common in robotic manipulation tasks, yet simulating them with both visual realism and physical accuracy remains challenging. Existing visual simulation methods typically rely on procedural geometric primitives that lack physically grounded deformation behavior, while physics-based approaches with robot learning support often approximate DLOs as rigid-link chains or generic soft bodies, failing to accurately capture the bending, twisting, and shear mechanics of slender elastic structures. In this work, we introduce DeformX, a co-simulation framework that integrates a dedicated Cosserat rod physics engine with NVIDIA Isaac Sim, enabling DLO simulations that are both physically faithful and visually realistic. Our Cosserat rod engine simulates the dynamics and self-collisions of DLOs, and contact interactions with arbitrary free-form meshes. To achieve high-fidelity visualization, we employ mesh skinning to map discrete rod deformations onto imported CAD models. To the best of our knowledge, DeformX is the one of the first frameworks for DLO simulation that unifies realistic visualization, principled physics, and compatibility with robot learning pipelines. We demonstrate its versatility across synthetic data generation and policy learning for DLO manipulation, and validate visual and physical fidelity through comparisons against real-world experiments. Notably, fine-tuning Segment Anything Model 3 (SAM3) on DeformX-generated data yields a 10.2\% mAP@75 improvement in real-image wire segmentation, and a rope-swinging policy trained entirely in DeformX achieves a mean target-hitting error of 6.6 cm on a UR5e manipulator in real-world trials, highlighting its strong sim-to-real transfer capability.

\end{abstract}

\section{INTRODUCTION}

Deformable linear objects (DLOs), including wires, cables, and ropes, are pervasive in robotics and industrial settings. They appear in applications such as wire harness assembly, cable routing, infrastructure inspection, and textile manipulation~\cite{intro_survey}. Accurate DLO simulation is therefore important for perception, planning, and control: a realistic simulator can enable scalable synthetic data generation for visual learning, support robot learning for manipulation policies, and reduce dependence on expensive and time-consuming real-world experimentation. Despite growing interest in DLO manipulation, existing approaches rarely satisfy three requirements simultaneously: (i) \emph{visual realism} for perception, (ii) \emph{physical fidelity} under gravity, contact, and manipulation, and (iii) \emph{compatibility with robot learning frameworks} for closed-loop training and evaluation. Most prior systems emphasize either visually plausible rendering with simplified dynamics or physically motivated models without an integrated, photorealistic robotics stack, which limits their utility for end-to-end perception and robot learning.

Existing approaches to visually simulating DLOs primarily rely on procedural modeling tools such as Blender, where DLOs are generated as random B\'ezier curves or as chains of simple geometric primitives like connected cylinders~\cite{handloom, wiredata}. These methods can produce visually plausible cable shapes under controlled conditions and are often used to generate synthetic datasets for training perception models~\cite{handloom,FASTDLO,wiredata,Zanella2021AutogeneratedWD,ISCUTE}, given the difficulty of collecting and annotating large-scale real-world wire data. For example, HANDLOOM~\cite{handloom} synthesizes 30{,}000 cable images for visual learning, but represents cables as random B\'ezier curves without physically grounded deformation and restricts appearance (e.g., uniformly white cables), limiting both mechanical realism and visual diversity. Similarly, \cite{wiredata} constructs synthetic cables as connected cylinders and uses collision modifiers to approximate interactions, which increases appearance variation but still relies on highly simplified mechanics. These procedural representations are typically not mesh-based, which prevents direct import of realistic cable CAD assets and reduces geometric fidelity and asset reusability. In addition, because the generated shapes are not governed by physically meaningful rod models, they may appear plausible in static renders yet fail to produce consistent dynamics under contact and self-interaction, conditions that matter for robot manipulation.

% On the visual side, large-scale real-world datasets for DLOs perception remain scarce due to the challenges of capturing and annotating thin, deformable objects in clutter. As a result, synthetic data generation has become a common substitute~\cite{handloom,FASTDLO,wiredata,Zanella2021AutogeneratedWD,ISCUTE}. Many pipelines rely on procedural modeling in tools such as Blender, where cables are instantiated as randomly sampled B\'ezier curves or as chains of simple primitives (e.g., connected cylinders)~\cite{handloom, wiredata}. 

On the physics side, common DLO simulators approximate deformable objects as chains of rigid capsules connected by ball joints~\cite{DexDLO, irp, R2S2R}. While computationally convenient and widely used in manipulation studies, rigid-link models oversimplify continuum mechanics (e.g., bending--twisting coupling and shear deformation), and often depends on extensive manual tuning of joint stiffness/damping parameters rather than directly interpretable material properties. An alternative is to model DLOs as volumetric soft bodies within general deformable-body frameworks, including finite-element and related methods~\cite{kaufmann2009flexible, wang2013efficient, lv2017physically, valentini2011modeling}. These approaches can express richer material behavior, but are typically inefficient for representing long, slender objects: key rod mechanics arise only implicitly through 3D discretization, computational cost can scale poorly with resolution, and results can be sensitive to meshing choices. In contrast, Cosserat rod theory provides a principled formulation specialized for DLOs by modeling a DLO as a one-dimensional continuum that explicitly captures stretching, shearing, bending, and twisting~\cite{cosserat_theory, linn2017discrete, cosserat_multisection}. Its dynamics are governed by physically meaningful parameters (e.g., Young's modulus and shear modulus), establishing a clearer link between simulation behavior and real material properties and enabling more accurate and interpretable modeling of DLOs~\cite{cable_modeling_review}. However, existing Cosserat rod solvers are often used in isolation from photorealistic simulators and robotics middleware. For example, libraries such as PyElastica~\cite{pyelastica_paper} provide physically realistic rod dynamics, but lack native support for complex interactions with free-form meshes and do not directly provide a high-fidelity rendering and robotics stack, limiting their applicability in robotic manipulation.

These limitations become especially consequential for robot learning with DLOs. DLO tasks remain bottlenecked by data acquisition and sim-to-real discrepancies~\cite{R2S2R, routing}. Many systems still rely heavily on real-world demonstrations because collecting reliable simulated rollouts with realistic dynamics and contacts is difficult. For instance, \cite{routing} reports collecting 1{,}442 teleoperated demonstration trajectories for cable routing, underscoring the cost of scaling real-world data. Some works attempt to train DLO manipulation policies purely in simulation, such as Iterative Residual Policy (IRP)~\cite{irp}, but their linked-capsule dynamics necessarily simplify elastic behavior, which can widen the sim-to-real gap and push algorithmic burden onto policy design rather than the simulator. Overall, there remains a gap in simulation infrastructure: a unified framework that jointly delivers physically grounded rod dynamics, visually realistic rendering with reusable CAD assets, and seamless integration with robot learning pipelines.

To address this gap, we propose \textbf{DeformX}, a co-simulation framework that couples a discrete Cosserat rod engine with Isaac Sim to achieve both physically and visually realistic DLO simulation for perception and robot learning. DeformX combines (i) physically interpretable Cosserat rod dynamics, (ii) realistic interactions with complex environments via mesh-based contact, and (iii) high-fidelity visualization and sensor simulation through CAD-compatible mesh skinning and Isaac Sim rendering. Our main contributions are:

\begin{itemize}
    \item \textbf{DeformX Framework:} A co-simulation system integrating a Cosserat rod physics engine with Isaac Sim to produce visually and physically realistic DLO simulations suitable for perception and robot learning.
    \item \textbf{Free-Form Mesh Contact:} Support for DLO interactions with free-form meshes within the Cosserat rod engine, enabling realistic rod-object contact in cluttered environments.
    \item \textbf{Mesh-Skinned Visualization with CAD Support:} We propose to utilize mesh skinning to map discrete Cosserat rod deformations onto meshes, allowing import of DLO CAD assets for high-fidelity visualization.
    \item \textbf{WireSeg-36k Dataset:} A synthetic wire instance segmentation dataset containing diverse RGB images, depth maps, and ground-truth annotations. Fine-tuning SAM3 on our dataset yields a 10.2\% improvement in mAP@75 on real-world experiments.
\end{itemize}

\setcounter{figure}{1}

\begin{figure}[h]
    \centering
    \includegraphics[width=\linewidth]{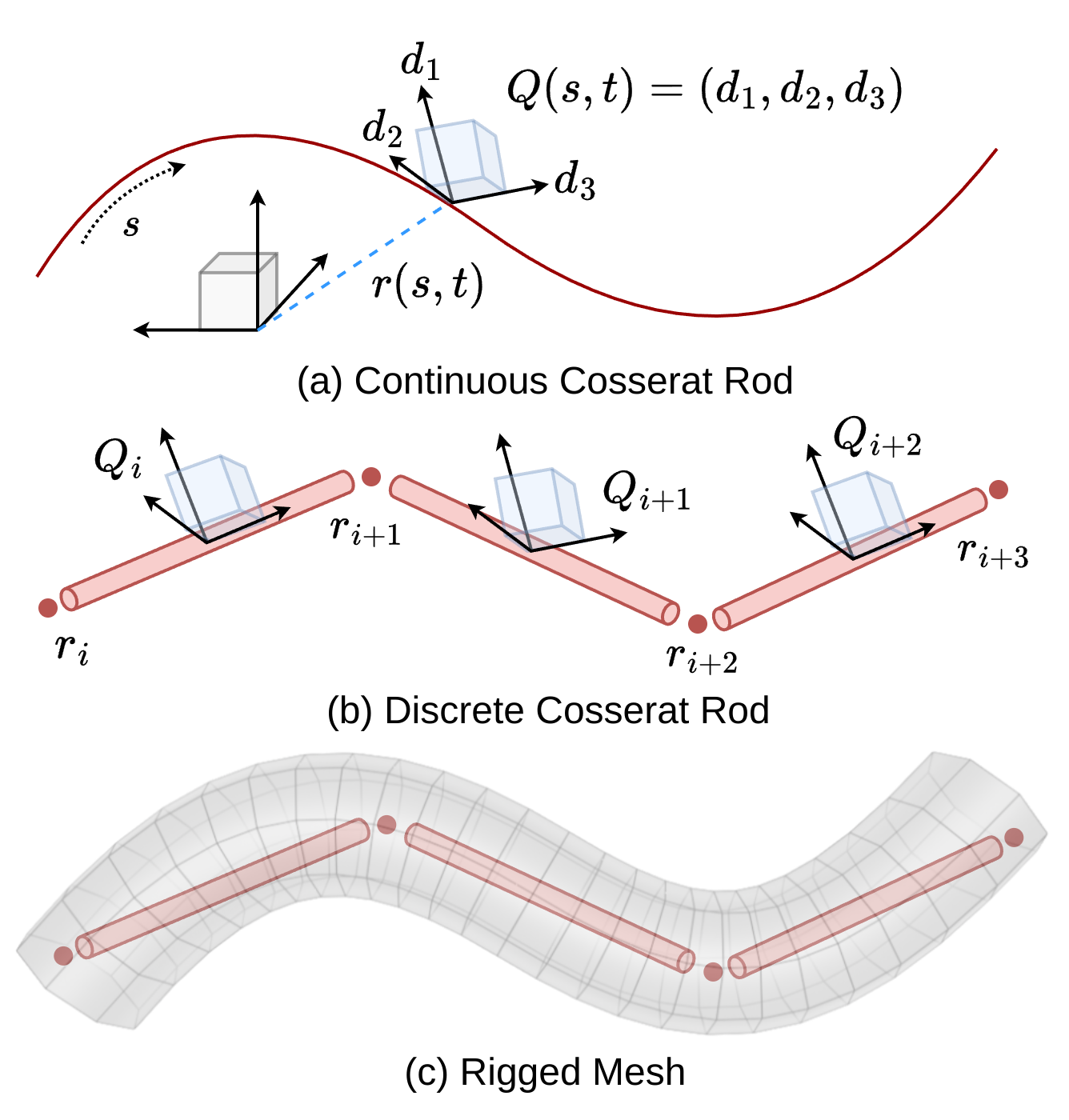}
    \caption{Cosserat rod modeling of DLOs. (a) Continuous Cosserat rod representation; (b) Discrete Cosserat rod representation; (c) Mesh skinned to the discrete Cosserat rod for realistic visualization.}
    \label{fig:notation}
\end{figure}

\begin{figure*}[t]
    \centering
    \includegraphics[width=1\linewidth]{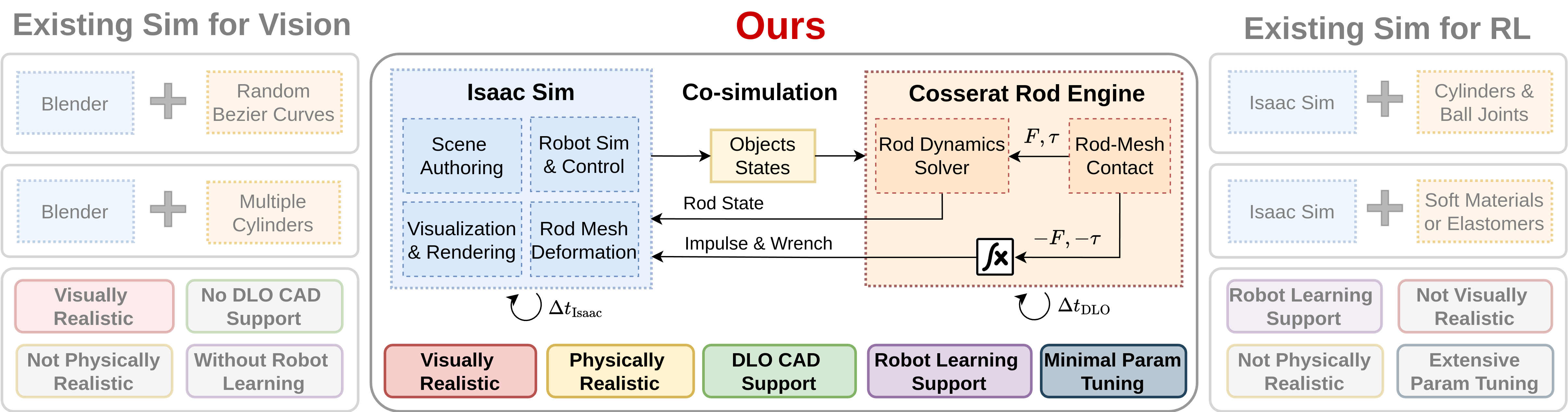}
    \caption{Overview of our co-simulation framework. Compared to existing simulators for vision (left) and RL (right), our framework jointly achieves visual realism, physical accuracy, DLO CAD support, and robot learning support with minimal parameter tuning by combining Isaac Sim and a dedicated Cosserat rod engine.}
    \label{fig:system_overview}
\end{figure*}

\section{NOTATIONS AND PRELIMINARY}

We model a deformable linear object (DLO) as a slender elastic rod using the Cosserat rod theory \cite{nonlinear_elas, pyelastica_notation}, which captures all deformation modes of a 1D continuum including stretching, shearing, bending, and twisting.

In the continuous representation (Fig.~\ref{fig:notation}a), the rod is parameterized by arc-length $s$ and time $t$, with centerline $\mathbf{r}(s,t)\in\mathbb{R}^3$ and an orthonormal material frame $\mathbf{Q}(s,t)=\{\mathbf{d}_1(s,t),\mathbf{d}_2(s,t),\mathbf{d}_3(s,t)\}\in \operatorname{SO}(3)$ attached to each cross-section, where the directors $\mathbf{d}_1,\mathbf{d}_2,\mathbf{d}_3$ are the columns of $\mathbf{Q}$.

In the discrete representation (Fig.~\ref{fig:notation}b), the centerline is sampled into vertices $\{\mathbf{r}_i(t)\}_{i=0}^{n}$ and orientations are stored as segment frames $\{\mathbf{Q}_j(t)\}_{j=0}^{n-1}$ associated with edges $\boldsymbol{\ell}_j=\mathbf{r}_{j+1}-\mathbf{r}_j$, yielding a finite-dimensional state $\{\mathbf{r}_i,\mathbf{Q}_j\}$~\cite{pyelastica_notation, discrete_elas, pyelastica_web}.
Time evolution is advanced numerically by discretizing $t$ into steps $t^k=k\Delta t$, so all state variables are updated from $(\mathbf{r}_i^{k},\mathbf{Q}_j^{k})$ to $(\mathbf{r}_i^{k+1},\mathbf{Q}_j^{k+1})$ at each timestep $\Delta t$~\cite{pyelastica_paper}.

Cosserat rod dynamics are computed by conservation of linear and angular momentum along the rod.
Material properties are incorporated through linear elastic constitutive laws, where Young's modulus $E$ and shear modulus $G$ determine the rod stiffness in stretching/shearing and bending/twisting~\cite{cosserat_theory}. External interactions are incorporated through forcing and boundary-condition terms, including gravity, endpoint loads/torques, and contact with friction.

\section{METHODS}
\subsection{Framework Overview}
We propose a co-simulation framework that tightly integrates a Cosserat rod engine with Isaac Sim to achieve physically accurate and visually realistic simulation of DLOs. The core challenge lies in bridging two systems with fundamentally different functionalities and time scales. To address this, we design a multi-rate scheme that handles state updates, cross-engine interactions, and ensures stable bidirectional coupling between the Cosserat rod engine and Isaac Sim. The proposed framework enables simulation of DLOs that is both visually and physically realistic, facilitating downstream visual perception learning and robot learning tasks. An overview of our framework is illustrated in Fig.~\ref{fig:system_overview}.

\subsection{Co-simulation Framework}
\noindent\textbf{Division of Functionality:} Our framework partitions functionalities between the Cosserat rod engine and Isaac Sim while maintaining tightly coupled interaction. The Cosserat rod engine computes all DLO dynamics and contacts between DLOs and other objects, ensuring physically consistent deformation and contact responses, while Isaac Sim handles the rigid-body environment, including robot manipulators, fixtures, and other scene elements, providing rigid-body simulation, scene management, control interfaces, and photorealistic rendering. This division ensures that robot dynamics and non-DLO interactions remain within Isaac Sim's domain, while deformation physics and rod-driven contacts are governed by the specialized rod engine. As a result, our framework preserves high physical fidelity for deformable objects while leveraging the robustness and scalability of a mature rigid-body simulation platform.

\begin{algorithm}[t]\RaggedRight
\caption{Multi-rate co-simulation}
\label{alg:cosim}
\begin{algorithmic}[1]
\Require Macro step $\Delta t_{\mathrm{Isaac}}$, substeps $N$, with $\Delta t_{\mathrm{DLO}}=\Delta t_{\mathrm{Isaac}}/N$
\For{each Isaac step $k$}
    \State Get objects states from Isaac Sim
    \Statex \hspace{\algorithmicindent} (pose/velocity/acceleration)
    \For{$n=1$ to $N$}
        \State Estimate motions of interacting objects
        \State Integrate DLO dynamics for $\Delta t_{\mathrm{DLO}}$
        \Statex \hspace{\algorithmicindent} (Cosserat rod + contact)
    \EndFor
    \State Export DLO skeleton poses for rendering
    \State Return reaction wrenches/impulses to Isaac Sim
    \State Isaac Sim advances rigid-body + control for $\Delta t_{\mathrm{Isaac}}$
\EndFor
\end{algorithmic}
\end{algorithm}

\noindent\textbf{Multi-rate Co-simulation:} A fundamental challenge in our co-simulation framework stems from the mismatch in time scales between the Cosserat rod engine and Isaac Sim modules. Stable DLO dynamics computation typically necessitates finer time steps, $\Delta t_{\mathrm{DLO}} \sim 10^{-5}$ s, whereas Isaac Sim commonly advances rigid-body dynamics and control at coarser time steps, $\Delta t_{\mathrm{Isaac}} \sim 10^{-2}$ s. Consequently, the DLO states are updated multiple times within a single Isaac Sim step, during which no direct interaction between these two modules is available.

To address this issue, we replicate the semi-implicit Euler integration used by Isaac Sim PhysX within the Cosserat rod engine. This approach enables us to estimate the motion of interacting rigid bodies (e.g., robot grippers or end-effectors) during fine-grained DLO updates, using only the states provided by Isaac Sim at coarse-grained time steps even when intermediate inputs from Isaac Sim are unavailable.

Specifically, we represent the pose of an interacting rigid body as a rigid transform $T\in SE(3)$ and parameterize incremental motion using a 6D minimal coordinate vector $\boldsymbol{\xi}\in\mathbb{R}^6$, where $\boldsymbol{\xi}=[\mathbf{x};\boldsymbol{\theta}]$ stacks translation $\mathbf{x}\in\mathbb{R}^3$
and a 3D rotation coordinate $\boldsymbol{\theta}\in\mathbb{R}^3\cong\mathfrak{so}(3)$. The corresponding generalized velocity and acceleration are denoted by $\dot{\boldsymbol{\xi}}$ and $\ddot{\boldsymbol{\xi}}$, respectively. At each time step $\Delta t$, we first update the generalized velocity using semi-implicit Euler:
\begin{equation}
\dot{\boldsymbol{\xi}}_{n+1}=\dot{\boldsymbol{\xi}}_n+\ddot{\boldsymbol{\xi}}_n\Delta t,
\end{equation}
and then update the pose by composing the current transform with the incremental motion on $\operatorname{SE}(3)$ induced by the updated velocity:
\begin{equation}
    T_{n+1}=T_n\Exp(\dot{\boldsymbol{\xi}}_{n+1}\Delta t),
\end{equation}
where $\Exp(\cdot)$ denotes the exponential map that converts a 6D motion vector in $\mathbb{R}^6\cong\mathfrak{se}(3)$ into an element of $\operatorname{SE}(3)$. 

By replicating Isaac Sim's integration scheme within the rod engine, we ensure consistent state evolution across simulation modules during multi-rate updates, while allowing contact interactions to be evaluated against continuously updated rigid-body motion at the DLO time scale. These contact interactions must also be propagated back to Isaac Sim. To this end, we accumulate contact forces computed at the DLO time scale into integrated impulses or wrenches, which are then fed back to the corresponding rigid bodies in Isaac Sim at each coarse simulation step. This strategy enables stable and physically consistent bidirectional coupling across disparate time resolutions.

\noindent\textbf{Modular Interface:} We have implemented the Cosserat rod engine as a Python module that integrates directly into Isaac Sim’s scripting environment, supporting both interactive UI-based workflows and headless execution. Users can construct scenes within Isaac Sim, after which geometry, poses, and simulation parameters are exported as initial conditions for the rod engine. The same interface is reused for large-scale dataset generation and robot learning, providing a unified pipeline from scene authoring to physics simulation and photorealistic rendering.

\subsection{Interaction with Free-Form Meshes}

For DLO simulation in robotic settings, interaction with surrounding objects is essential. However, current PyElastica and other Cosserat rod libraries~\cite{pyelastica_repo, pyelastica_paper, stable_rod} lack native support for contact with free-form meshes. Building on PyElastica's penalty-based contact formulation, we compute the closest-point distance $d$ between rod nodes $\{\mathbf{r}_i\}$ and the mesh surface, and apply a repulsive normal contact force parameterized by stiffness $k$ and dissipation $\nu$. The Cosserat rod engine applies the resulting contact forces to the rod internally, and returns the equal-and-opposite reaction impulses/wrenches to the mesh, where the corresponding rigid body dynamics are resolved \cite{pyelastica_repo}.

To improve computational efficiency, we prebuild a Bounding Volume Hierarchy (BVH) over the mesh and employ AABB-based broad-phase pruning to reduce distance queries. To enhance robustness under larger time steps, we introduce a repulsion distance for watertight meshes to prevent deep penetration and keep the rod outside the surface.

\subsection{DLO Visualization in Isaac Sim}
For realistic visualization in Isaac Sim, we represent each DLO with a smooth tubular mesh that is skinned to the discrete Cosserat rod vertices. In this way, the motion of the Cosserat rod smoothly deforms the mesh. During simulation, the rod dynamics solver outputs the rod centerline and local orientation for each element, and at every Isaac Sim step we update the tubular mesh accordingly using these rod states.

Mesh skinning is a technique widely used in animation and character modeling~\cite{james2005skinning}. We incorporate this technique with a discretized Cosserat rod model to achieve high visual fidelity in Isaac Sim while maintaining full consistency with the underlying rod dynamics, as illustrated in (Fig.~\ref{fig:notation}c). Moreover, because DLOs are visualized using skinned meshes, we can directly import CAD models of DLOs with pre-defined skins. This allows us to leverage a wide range of existing DLO CAD assets to create diverse and highly realistic DLO representations.

\section{Experiment}
\subsection{Validation of DLO Physics Simulation}
To validate physical fidelity, we perform a set of physics sanity checks and quantitative comparisons against observed DLO motion. Unlike Isaac Sim's built-in DLO approximations, the Cosserat rod model explicitly captures stretching, shearing, bending, and twisting, enabling characteristic behaviors such as gravity-induced equilibria, curvature propagation, and torsion-bending coupling.
We validate the physical fidelity of our simulation through two real-world experiments:
% Crucially, our parameters have clear physical meaning (e.g., density, radius, Young's modules and shear modules, bending/torsional stiffness, and damping), which supports calibration to real DLOs and principled material variation rather than heuristic tuning. This makes the simulator well-suited for data generation and for analyzing sim-to-real sensitivity in DLO manipulation.

\begin{figure}
    \vspace{5pt}
    \centering
    \includegraphics[width=\linewidth]{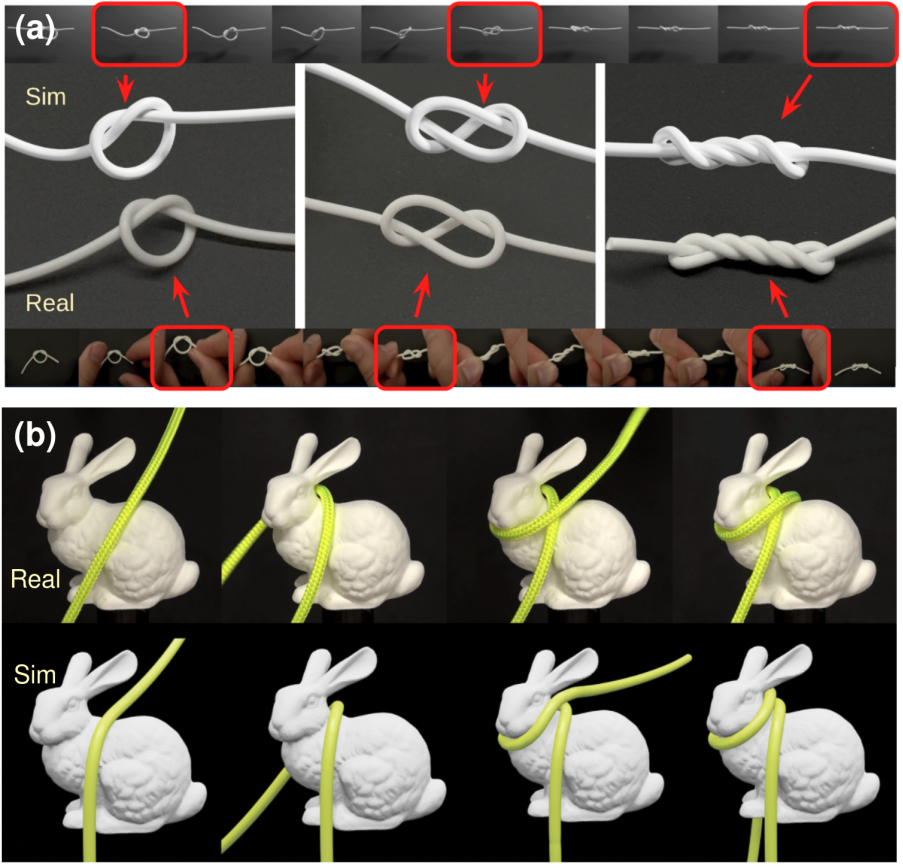}
    \caption{Sim2real comparison for DLOs. (a) We tie a trefoil knot on a wire, apply continuous twists to induce shape transitions, and qualitatively compare the representative configurations between simulation (top) and real experiments (bottom) \cite{discrete_elas}. (b) shows flexible rope (scarf) wrapping around a bunny with multi-point contact, sliding, and self-contact.}
    \vspace{-10pt}
    \label{fig:demo1}
\end{figure}

\textbf{(1) Knot deformation under twist:}
We design the first experiment to assess the simulator's ability to reproduce realistic quasi-static deformations given real-world measurements. The physics-based parameters in our simulator are computed from factory-provided material specifications\cite{tpu}.
We tie the DLO into a simple (trefoil) knot and apply controlled twisting \cite{discrete_elas}. As shown in Fig.~\ref{fig:demo1}(a), the simulated knot undergoes the same qualitative shape transitions as those observed in real world.

\textbf{(2) Free-form contact modeling:}
We evaluate our implementation of contact handling by wrapping a flexible rope (scarf) around a rigid bunny model. Starting from an initially draped configuration, we progressively pull and slide the scarf to induce multi-point, sliding contact and self-contact. Fig.~\ref{fig:demo1} visualizes representative snapshots, showing that the simulated scarf conforms to the bunny geometry, preserves stable contact without interpenetration, and produces realistic winding and settling behavior under gravity and friction.

For both experiments, our simulator yields physically consistent DLO behavior, supporting its use for realistic data generation and sim-to-real analysis.

\begin{figure}[h]
    \vspace{5pt}
    \centering
    \includegraphics[width=\linewidth]{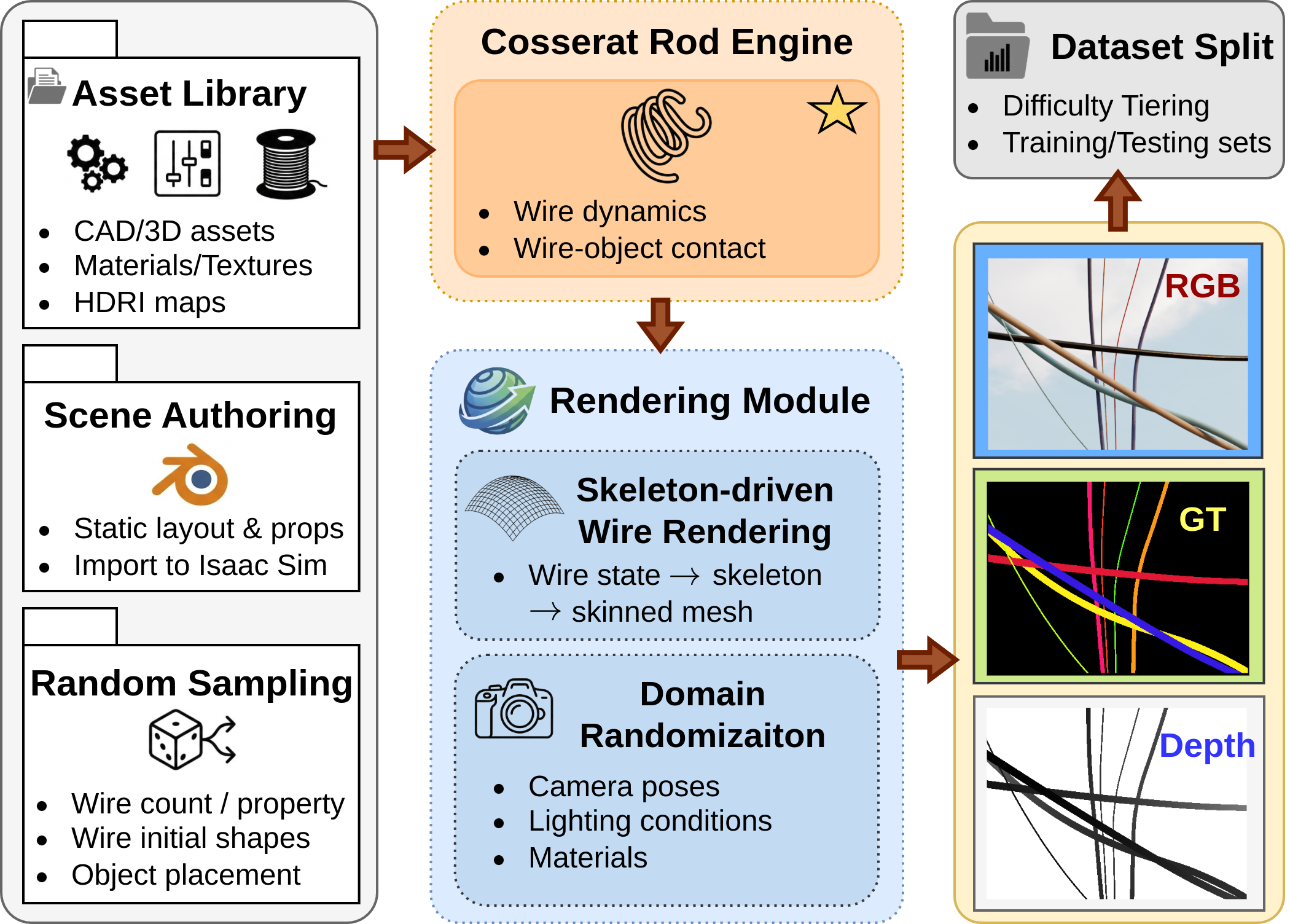}
    \caption{Synthetic wire segmentation dataset generation pipeline. Randomized scenes are built from asset libraries. Wires are simulated with a Cosserat rod engine, and visualized as skinned meshes. RGB images, segmentation masks, and depth maps are generated and split by difficulty levels.}
    \label{fig:data_pipeline}
\end{figure}

% Put once in your preamble (amssymb already loaded)
\newcommand{\Yes}{\checkmark}
\newcommand{\No}{\ensuremath{\times}}

\begin{table}[t]
\centering
\scriptsize
\setlength{\tabcolsep}{3pt}
\renewcommand{\arraystretch}{1.0}
\begin{tabular}{@{}lccccc@{}}
\toprule
Dataset &
\begin{tabular}[c]{@{}c@{}}Physics\\Realism\end{tabular} &
\begin{tabular}[c]{@{}c@{}}Instance\\Label\end{tabular} &
\begin{tabular}[c]{@{}c@{}}CAD\\Support\end{tabular} &
\begin{tabular}[c]{@{}c@{}}Visually-grounded\\Scenes\end{tabular} &
\begin{tabular}[c]{@{}c@{}}Num.\\Images\end{tabular} \\
\midrule
HANDLOOM \cite{handloom} & \No & \No & \No & \No & 30k \\
FASTDLO \cite{FASTDLO} & \No & \Yes & \No & \No & 32k \\
Fresnillo et al.\ \cite{wiredata} & \Yes & \No & \No & \Yes & 25k \\
Zanella et al.\ \cite{Zanella2021AutogeneratedWD} & \No & \No & \No & \No & 28.5k \\
ISCUTE \cite{ISCUTE} & \No & \Yes & \Yes & \Yes & 28k \\
\textbf{WireSeg-36k (Ours)} & \Yes & \Yes & \Yes & \Yes & 36k \\
\bottomrule
\end{tabular}
% \caption{Binary comparison between existing DLO datasets and our WireSeg-32k dataset. ``Physics realism'' = physics-based simulation of DLO deformation during data generation (e.g., falling/wind/collisions). ``Instance label'' = per-object instance masks (trace-only supervision counted as \No). ``CAD Support'' = able to represent DLO as a free-form mesh. ``Visually-grounded Scenes'' = image background is } 
\caption{Binary comparison between existing DLO datasets and our WireSeg-36k dataset. ``Physics realism'' indicates whether data generation uses physics-based simulation of DLO deformation (e.g., gravity, wind, and collisions). ``Instance label'' indicates whether per-object instance masks are provided (trace-only supervision is counted as \No). ``CAD support'' indicates whether DLOs can be represented as free-form meshes (enabling direct CAD import). ``Visually grounded scenes'' indicates whether cables are rendered within realistic, image-based backgrounds rather than on plain or purely synthetic backdrops.}
\label{tab:wire_dataset_binary}
\vspace{-5pt}
\end{table}

% \begin{table*}[t]
% \vspace{5pt}
% \centering
% \resizebox{\textwidth}{!}{%
% \setlength{\tabcolsep}{6pt}
% \renewcommand{\arraystretch}{1.15}
% \small
% \begin{tabular}{lcccccccccc}
% \toprule
% \multirow{2}{*}{\textbf{Model}} &
% \multicolumn{2}{c}{\textbf{Hard (Syn)}} &
% \multicolumn{2}{c}{\textbf{Medium (Syn)}} &
% \multicolumn{2}{c}{\textbf{Easy (Syn)}} &
% \multicolumn{2}{c}{\textbf{Total (Syn)}} &
% \multicolumn{2}{c}{\textbf{Total (Real)}} \\
% \cmidrule(lr){2-3}\cmidrule(lr){4-5}\cmidrule(lr){6-7}\cmidrule(lr){8-9}\cmidrule(lr){10-11}
% & \textbf{F1@75} & \textbf{mAP@75} &
% \textbf{F1@75} & \textbf{mAP@75} &
% \textbf{F1@75} & \textbf{mAP@75} &
% \textbf{F1@75} & \textbf{mAP@75} &
% \textbf{F1@75} & \textbf{mAP@75} \\
% \midrule
% SAM3 (Base)   & 0.169 & 0.052 & 0.454 & 0.291 & 0.818 & 0.783 & 0.456 & 0.328 & 0.296 & 0.157 \\
% SAM3 + LoRA   & 0.191 & 0.068 & 0.517 & 0.374 & 0.855 & 0.837 & 0.501 & 0.389 & 0.314 & 0.173 \\
% \midrule
% $\Delta$ (LoRA--Base) & +13.4\% & +31.1\% & +13.9\% & +28.8\% & +5.0\% & +6.8\% & +9.8\% & +18.4\% & +6.1\% & +10.2\% \\
% \bottomrule
% \end{tabular}}
% % \vspace{2pt}
% \caption{SAM3 performance with and without LoRA fine-tuning. For each subset, we report dataset-level F1@75 and COCO
% mAP@75 (higher is better) on the Easy/Medium/Hard tiers, the full synthetic set, and a held-out real test set. The last
% row reports the percent improvement from LoRA fine-tuning.}
% \vspace{-5pt}
% \label{tab:sam_result}
% \end{table*}

\begin{table*}[t]
\centering
\resizebox{\textwidth}{!}{%
\setlength{\tabcolsep}{4pt}\renewcommand{\arraystretch}{1.15}\small
\begin{tabular}{lccccccccccccccc}
\toprule
\multirow{2}{*}{\textbf{Model}} &
\multicolumn{3}{c}{\textbf{Hard (Syn)}} & \multicolumn{3}{c}{\textbf{Medium (Syn)}} &
\multicolumn{3}{c}{\textbf{Easy (Syn)}} & \multicolumn{3}{c}{\textbf{Total (Syn)}} &
\multicolumn{3}{c}{\textbf{Total (Real)}} \\
\cmidrule(lr){2-4}\cmidrule(lr){5-7}\cmidrule(lr){8-10}\cmidrule(lr){11-13}\cmidrule(lr){14-16}
& \textbf{F1@75} & \textbf{mAP@75} & \textbf{$J$}
& \textbf{F1@75} & \textbf{mAP@75} & \textbf{$J$}
& \textbf{F1@75} & \textbf{mAP@75} & \textbf{$J$}
& \textbf{F1@75} & \textbf{mAP@75} & \textbf{$J$}
& \textbf{F1@75} & \textbf{mAP@75} & \textbf{$J$} \\
\midrule
SAM3 (Base) & 0.169 & 0.052 & 0.467 & 0.454 & 0.291 & 0.676 & 0.818 & 0.783 & 0.848 & 0.456 & 0.328 & 0.679 & 0.296 & 0.157 & 0.486 \\
SAM3 + LoRA & 0.191 & 0.068 & 0.518 & 0.517 & 0.374 & 0.738 & 0.855 & 0.837 & 0.879 & 0.501 & 0.389 & 0.728 & 0.314 & 0.173 & 0.496 \\
\midrule
$\Delta$ (LoRA--Base) & +13.4\% & +31.1\% & +10.9\% & +13.9\% & +28.8\% & +9.2\% & +4.5\% & +6.8\% & +3.7\% & +9.8\% & +18.4\% & +7.2\% & +6.1\% & +10.2\% & +2.1\% \\
\bottomrule
\end{tabular}}
\caption{SAM3 with and without LoRA fine-tuning. Dataset-level F1@75, COCO mAP@75, and mean per-image Jaccard $J$ on the Easy/Medium/Hard synthetic tiers, the full synthetic set, and a held-out real test set. Last row: \% improvement from LoRA.}
\label{tab:sam_result}                           
\end{table*}

\subsection{Applications: Data Generation}
\noindent\textbf{Overview:} A key application of a visually and physically realistic DLO simulator is to generate high-quality synthetic data for training perception models without expensive manual annotation. Wire instance segmentation is fundamental to many tasks, yet large-scale annotated datasets are scarce due to the difficulty of labeling thin, deformable objects in cluttered scenes. With our co-simulation framework, we construct \textit{WireSeg-36k}, a wire instance segmentation dataset designed with three principles: (1) high diversity in configurations, (2) a broad spectrum of difficulty levels for targeted training and evaluation, and (3) scenario-specific categories aligned with real-world use cases. A comparison between our dataset and existing datasets is shown in Table~\ref{tab:wire_dataset_binary}.

\begin{figure}[!t]
    \centering
    \includegraphics[width=\linewidth]{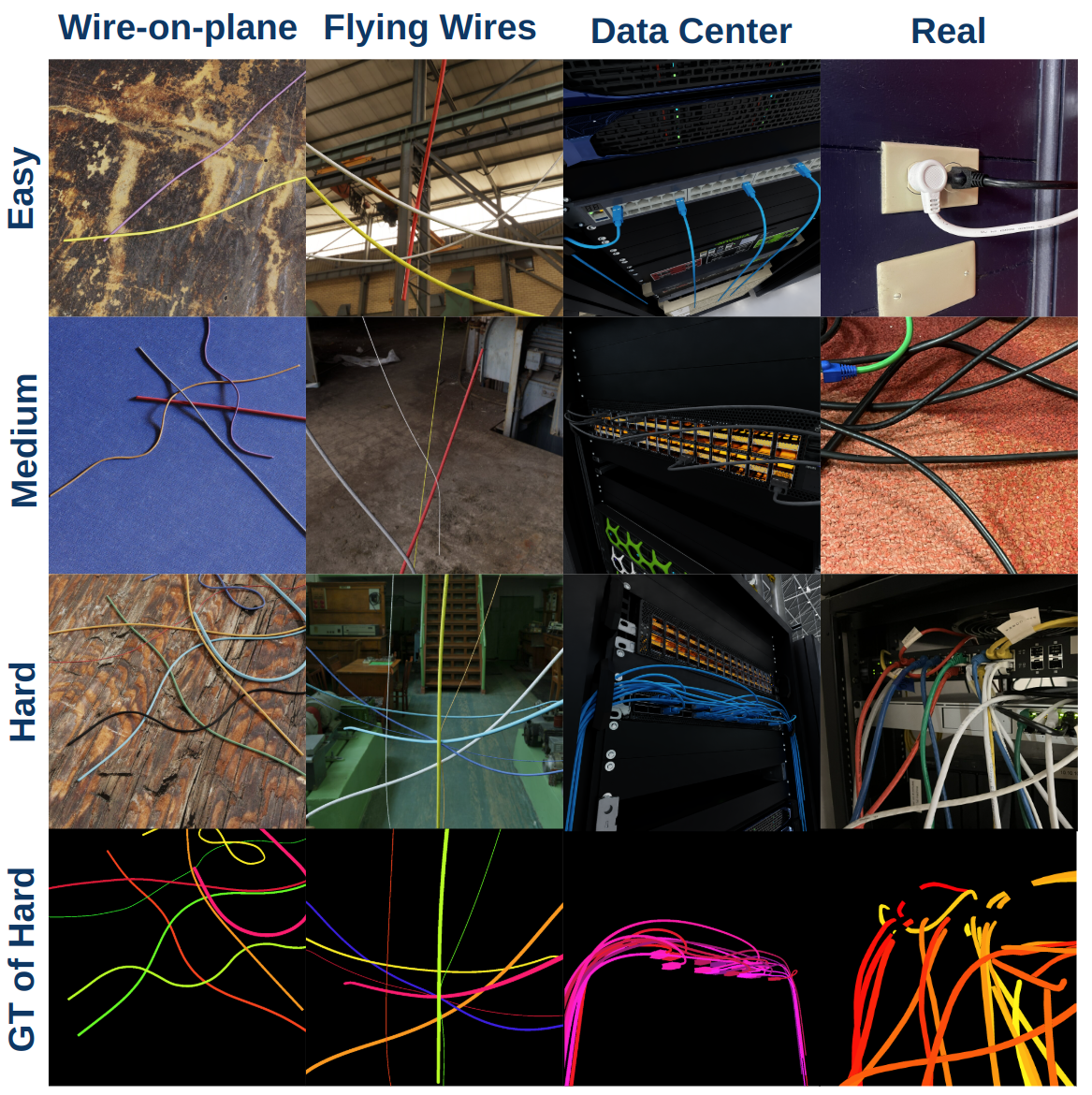}
    \caption{Representative images from three settings across Easy/Medium/Hard, with Hard ground-truth instance masks.}
    \label{fig:dataset}
\end{figure}

\noindent\textbf{Data Generation Pipeline:} The data generation pipeline is illustrated in Fig.~\ref{fig:data_pipeline}. We assemble a library of base scenes from publicly available 3D assets and then import the resulting geometry and materials into Isaac Sim. For each scene instance, we procedurally generate wire configurations (e.g., number of wires, length, radius, and initial shapes) and sample rigid objects/assets and their placements in Isaac Sim. During data generation, Isaac Sim simulates the robot and all non-wire rigid bodies, while our Cosserat Rod Engine solves the wire dynamics and computes wire-object interaction forces that are exchanged with Isaac Sim through the coupling interface. The simulated wire state is streamed to Isaac Sim for skeleton-driven rendering, together with the static scene geometry and asset materials. We further randomize camera viewpoints, lighting conditions, and scene textures to increase visual diversity and improve robustness to domain shifts. Sample images from our dataset is shown in Fig.~\ref{fig:dataset}.

\noindent\textbf{Categories:} We organize the dataset into three scenario categories: (1) \textbf{wire-on-plane}, a canonical tabletop setup; (2) \textbf{flying wires}, focusing on gravity-driven dangling dynamics; and (3) \textbf{data center}, featuring cables arranged within data-center racks, using photorealistic assets from NVIDIA Omniverse\cite{data_center}. In total, we collect 36,000 rendered images from 720 independent simulation runs.

\noindent\textbf{Annotations, Depth, and Difficulty:}
We provide per-wire instance masks for multi-wire segmentation, along with per-pixel depth maps to support high-resolution 3D perception beyond typical RGB-D limitations on thin cables~\cite{trackDLO,depth_anything_3}. We also stratify the dataset into easy/medium/hard splits by thresholding the per-image Jaccard score $J$ (mask IoU) achieved by the off-the-shelf SAM3 baseline on the generated images (Fig.~\ref{fig:dataset}). Images with $J < 0.6$ are labeled hard, those with $J \geq 0.8$ easy, and all others medium.

% \subsection{Ground-truth Labels}
% To support instance-level segmentation and tracking of multiple wires, we provide per-wire masks for every image, which is uncommon in existing wire segmentation datasets.

% \subsection{Depth Labels}
% Depth cues are also critical for wire manipulation. While many prior systems rely on RGB-D sensors\cite{trackDLO}, their depth maps are often low-resolution and unreliable for thin cables. Motivated by recent progress in learning-based depth estimation \cite{depth_anything_3}, we provide a per-pixel depth map for each image. These depth labels can be used to train wire-specific depth models, enabling high-resolution 3D perception.

% \subsection{Difficulty Assessment}
% To support different training objectives and robustness evaluations, we partition the dataset into easy, medium, and hard subsets. We estimate difficulty by running SAM3 on the generated images, ranking samples by accuracy, and splitting them into three tiers, as shown in Fig.~\ref{fig:dataset}. Qualitatively, we find that illumination, background complexity, wire diameter, and wire layouts are among the dominant factors affecting difficulty.

\noindent\textbf{Complementary Real-World Test Set:} To support real-world evaluation, we additionally create a separate test set of 300 in-the-wild images with manually annotated ground truth segmentations. Images are collected across diverse backgrounds, lighting conditions, and cable types to reflect real deployment variation.

% \noindent\textbf{Model Fine-tuning:}
% We conducted a simple experiment of LoRA-fine tuning Segmentation Anything Model 3 (SAM3) to showcase our dataset can be used to produce vision models for wire instance segmentation. We insert LORA layers...

% We tune the vision encoder and mask decoder by inserting light-weight LoRA adapters into the attention projections (\texttt{q_proj}, \texttt{k_proj}, \texttt{v_proj}; rank 16, $\alpha$=32, dropout 0.05), while leaving other components frozen to keep the update parameter-efficient and stable. Training uses a conservative learning rate ($10^{-5}$), an effective batch size of 16, and 5 epochs on our synthetic wire dataset.

% We evaluate SAM3 on our dataset as a strong foundation-model baseline. We run SAM3 in text-prompt mode with the prompt \texttt{``cable''}, then post-process the output into per-wire instance masks using thresholding, connected components, and small-component filtering. Predicted instances are aligned to ground-truth wires by maximum-IoU matching. Some qualitative segmentation results are shown in Fig.~\ref{fig:SAM}, and we report our performance in Tab.~\ref{tab}

% The \textbf{instance-level F1@50} is reported, where a predicted wire is counted as a true positive if it matches a ground-truth instance with IoU $\ge 0.50$ (with one-to-one matching), and compute precision/recall and the resulting F1 score across the dataset. We summarize results by difficulty tier and on a held-out real-world test set, as shown in Table~\ref{tab:sam_result}.

\noindent\textbf{Model Fine-tuning:}
To demonstrate that our dataset can directly support training wire instance segmentation models, we perform a simple LoRA fine-tuning experiment on SAM3, using $1024\times1024$ inputs with $288\times288$ output masks.
We adapt the vision encoder and mask decoder by inserting lightweight LoRA adapters into the attention projections (\texttt{q\_proj}, \texttt{k\_proj}, \texttt{v\_proj}; rank~16, $\alpha=32$, dropout~0.05), while keeping all other components frozen to maintain parameter efficiency and training stability. We train for 5 epochs on our synthetic wire dataset using a learning rate of $1\times10^{-5}$ and an effective batch size of 16.

As a strong foundation-model baseline, we also evaluate off-the-shelf SAM3 on our dataset in text-prompt mode using the prompt \texttt{``cable''}. We report dataset-level F1@75 (instance F1 with one-to-one matching at IoU $\ge 0.75$) and mAP@75 (COCO-style AP at IoU $=0.75$). Across both the synthetic benchmark and our held-out real-world test set, LoRA fine-tuning improves performance over the off-the-shelf baseline (Table~\ref{tab:sam_result}).

\subsection{Applications: Robot Learning}

\noindent\textbf{Motivation:}
Dynamic rope swinging amplifies modeling errors: small inaccuracies in bending/torsion and contact can cause large deviations in the rope-tip trajectory, leading to a pronounced sim-to-real gap in fast ``whipping'' regimes. We therefore use a planar hit-target rope-swinging benchmark inspired by Chi \emph{et al.}~\cite{irp} as a stress test to quantify whether improved DLO physics yields more reliable sim-to-real policy transfer.

\noindent\textbf{Task and Metric:}
A UR5e robotic arm swings a rope so that its tip passes as close as possible to a fixed target point. Following~\cite{irp}, we evaluate performance using the minimum tip-to-goal distance over a rollout,
\begin{equation}
d_{\min} \;=\; \min_{t \in [0,T]} \left\| \mathbf{p}_{\mathrm{tip}}(t) - \mathbf{p}_{\mathrm{goal}} \right\|_2.
\label{eq:dmin}
\end{equation}
We parameterize the robot motion with a low-dimensional joint-space action
\begin{equation}
\vspace{-2pt}
a_t = (\Delta J_1, \Delta J_2, \Delta J_3),
\end{equation}
where $\Delta J_1$--$\Delta J_3$ denote bounded per-step increments for the three actuated UR5e joints. These increments are accumulated into joint position targets at each control step.

\noindent\textbf{Planar Motion Constraint:}
We constrain the robot motion to be planar, and define the target location in-plane. This restriction is sufficient for the hit-target task, since out-of-plane targets can be reached by a trivial rotation of the robot base; we therefore fix the base rotation and focus on planar whipping dynamics.

\noindent\textbf{Parameter Settings of Simulated Ropes:}
The choice of rope parameters can significantly affect experimental outcomes. To ensure a fair comparison between DeformX and the Isaac Sim linked-capsule baseline, we calibrate the rope parameters using a robot-driven rope motion experiment. Specifically, we attach a 2\,m rope to a UR5e robot arm and command a sinusoidal end-effector trajectory (1.2\,Hz frequency and 0.2\,m amplitude), while tracking the rope in 3D using a motion-capture system. The identical boundary motion is then replayed in simulation. We tune the rope parameters so that the trajectory error between simulated and real rope configurations is minimized for both simulators:
\begin{equation}
    \vspace{-3pt}
   Error(t)=\frac{1}{N}\sum_{i=1}^{N}\left\lVert \mathbf{x}^{\mathrm{sim}}_{i}(t)-\mathbf{x}^{\mathrm{real}}_{i}(t)\right\rVert_{2}, 
   \label{eq:error}
   \vspace{0pt}
\end{equation}

where \(N\) is the number of motion-capture markers placed along the rope (here \(N=20\)).

\begin{figure}[t]
    \vspace{5pt}
    \centering
    \includegraphics[width=\linewidth]{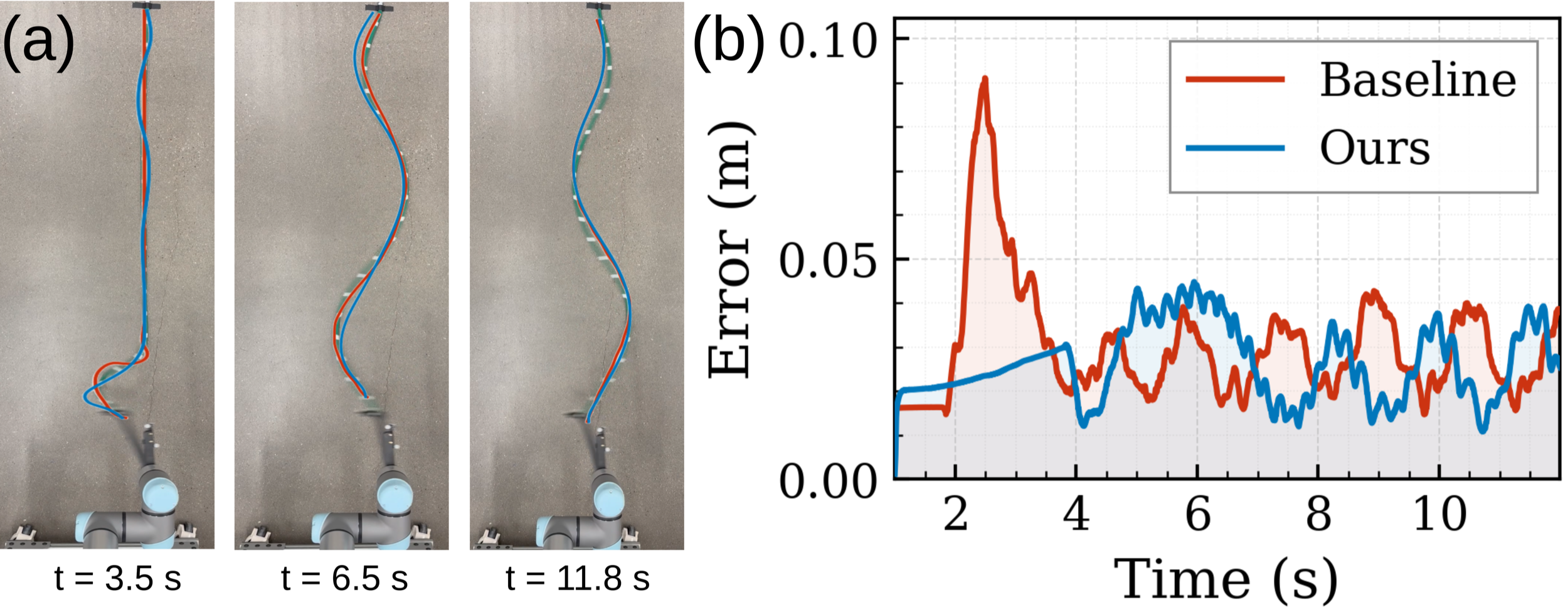}
    \caption{(a) Robot-driven rope experiment with one end anchored to the ground and the other attached to the robot end-effector, which executes a sinusoidal trajectory; overlays are shown with time stamps. (b) Trajectory error over time, as defined in Eq.~\ref{eq:error}.}
    \label{fig:sin_motion}
    \vspace{-10pt}
\end{figure}

The resulting trajectory errors are shown in Fig.~\ref{fig:sin_motion}. The calibrated rope parameters produce small and comparable errors for both simulators. These parameters are subsequently used for RL training.

\noindent\textbf{Controlled Comparison and Training:}
To isolate the effect of the DLO simulator, we keep the PPO algorithm, observations/actions, reward, training budget, and all non-DLO task components fixed, and vary only the DLO backend: (i) DeformX and (ii) an Isaac Sim linked-capsule baseline. PPO is trained with identical rollout, parallelization, and state/goal inputs for both backends. The policy controls a planar 3-DoF subset of UR5e joints via bounded incremental position commands, with the same action space and command limits for both backends.

\begin{figure}[t]
    \vspace{5pt}
    \centering
    \includegraphics[width=\linewidth]{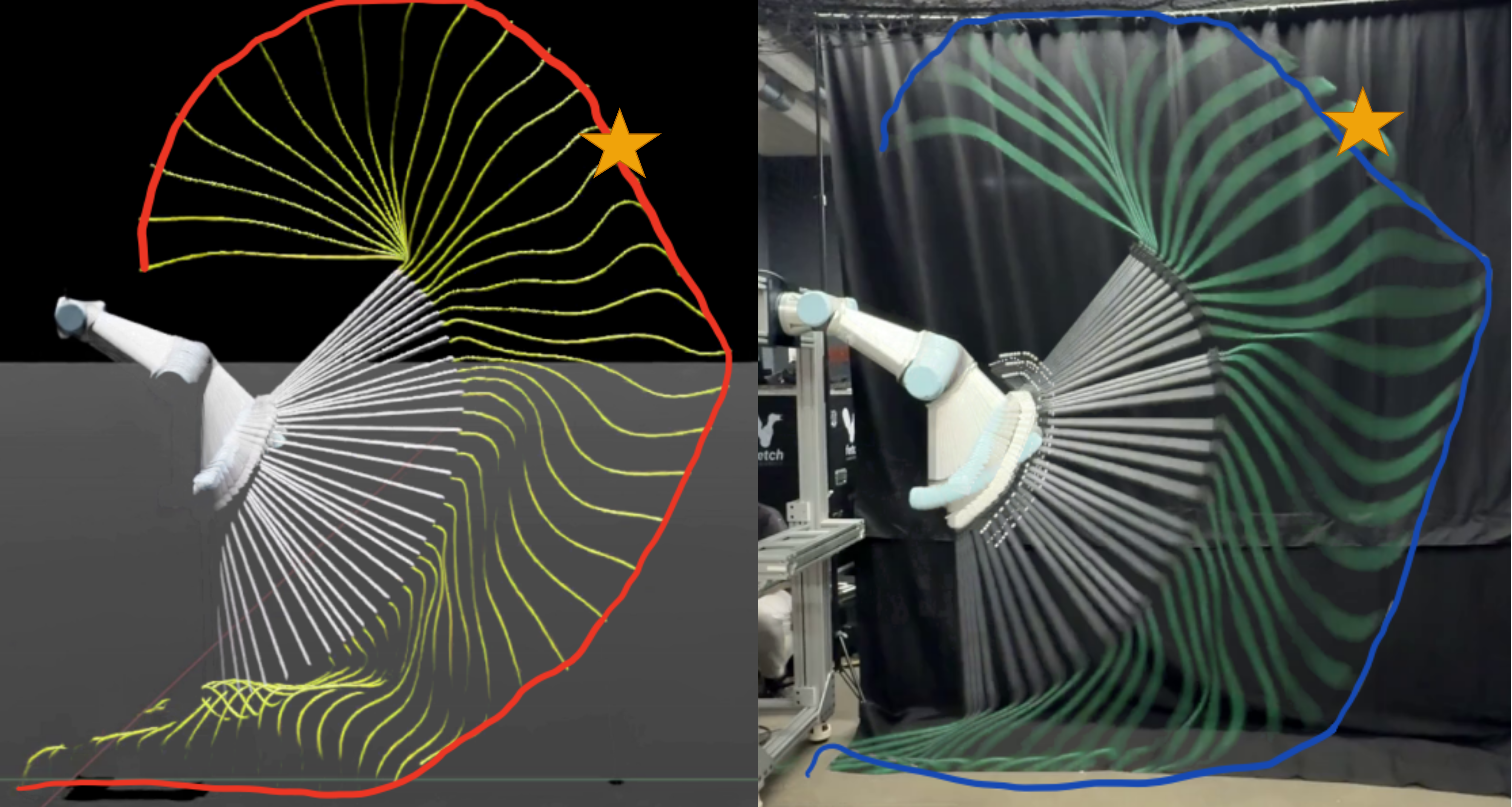}
    \caption{Goal-conditioned dynamic manipulation of a DLO. Our RL policy reaches the goal. Left: simulation rollout. Right: real-world rollout. The goal location is marked by a star. Colored curves visualize the DLO configuration over time during a rollout.}
    \label{fig:demo3}
\end{figure}

\noindent\textbf{Sim-to-real Evaluation:}
We deploy the learned open-loop motion on a real UR5e and track the rope tip using an OptiTrack system. For a fixed goal, we repeat the same execution $n{=}10$ times and report the mean and standard deviation of $d_{\min}$. Representative rollouts are shown in Fig.~\ref{fig:demo3}, and results are summarized in Table~\ref{tab:rl_hit_target}. While both simulators obtain low in-simulation errors, only DeformX transfers reliably to the real robot. Across all three targets, DeformX reduces the real-world minimum tip-to-goal distance from 15.1/25.9/30.4 cm to 6.6/7.3/5.8 cm, respectively. Since both simulators are calibrated on the same robot-driven rope experiment and all PPO settings are kept fixed, this result indicates that physically accurate modeling of bending, torsion, and contact is critical for sim-to-real transfer in dynamic rope swinging.

\begin{table}[t]
\centering
\small
\setlength{\tabcolsep}{3pt}
\begin{tabular}{lccc}
\toprule
\textbf{Target Point} &\textbf{Method} & \textbf{Sim} $d_{\min}$ & \textbf{Real ($n{=}10$)} $d_{\min}$ \\
\midrule
\multirow{2}{*}{(0, 200, 230)} & Baseline & 4.9 & $15.1 \pm 6.1$ \\ & \textbf{DeformX (Ours)} & 4.2 & $\textbf{6.6} \pm 4.7$ \\
\midrule
\multirow{2}{*}{(0, 200, 150)} & Baseline & 4.4 & $25.9 \pm 8.9$ \\ & \textbf{DeformX (Ours)} & 1.4 & $\textbf{7.3} \pm 1.2$ \\
\midrule
\multirow{2}{*}{(0, 170, 50)} & Baseline & 4.3 & $30.4 \pm 14.3$ \\ & \textbf{DeformX (Ours)} & 3.3 & $\textbf{5.8} \pm 3.2$ \\
\bottomrule
\end{tabular}
\caption{Hit-target task results. $d_{\min}$ (unit: cm) is defined in Eq.~\eqref{eq:dmin} and reported in centimeters. Real-world results are computed over $n{=}10$ repeated executions for the same goal; we report mean and standard deviation.}
\label{tab:rl_hit_target}
\end{table}

% \begin{table}[h]
% \centering
% \small
% \setlength{\tabcolsep}{5pt}
% \begin{tabular}{lcccc}
% \hline
% \textbf{Training Sim} & \textbf{Eval} & \textbf{Succ.} $\uparrow$ & $d_{\min}$ (cm) $\downarrow$ & $E_{\text{traj}}$ (cm) $\downarrow$ \\
% \hline
% Isaac rod proxy & Sim  & 0.00 & 0.0 & -- \\
% Ours (Cosserat) & Sim  & 0.00 & 0.0 & -- \\
% \hline
% Isaac rod proxy & Real & 0.00 & 0.0 & 0.0 \\
% Ours (Cosserat) & Real & 0.00 & 0.0 & 0.0 \\
% \hline
% \end{tabular}
% \caption{Hit-target task results (replicating~\cite{irp}). Succ.: success rate within tolerance $\epsilon$; $d_{\min}$: minimum tip-to-target distance over the rollout; $E_{\text{traj}}$: sim-to-real trajectory mismatch for the rope tip (mean position error over time).}
% \label{tab:rl_hit_target}
% \end{table}

\section{Conclusion and Discussion}

We introduced \textit{DeformX}, a co-simulation framework that integrates a dedicated Cosserat rod engine with NVIDIA Isaac Sim to enable DLO simulation that is both physically faithful and visually realistic, while remaining compatible with scalable data generation and robot learning. Our framework supports contact interactions with free-form meshes, multi-rate coupling with rigid-body dynamics, and CAD-quality visualization through mesh skinning. Leveraging \textit{DeformX}, we further constructed \textit{WireSeg-36k}, a wire instance segmentation dataset that provides RGB images, depth maps, and ground-truth annotations.

Despite these advantages, there are several limitations. First, our current Cosserat rod engine is implemented entirely on the CPU, which limits computational efficiency and scalability. Second, in the present multi-rate co-simulation scheme, the influence of the DLO on contacting rigid bodies is approximated by integrating contact forces into a single impulse or wrench within each Isaac Sim time step. This approximation may introduce inaccuracies in the bidirectional coupling between deformable and rigid objects. Additionally, recent advances in stable Cosserat rod formulations~\cite{stable_rod} propose a split position-rotation optimization scheme with a closed-form Gauss-Seidel quasi-static orientation update, achieving high accuracy under large stiffness parameters while maintaining stability with larger time steps. Such methods also support GPU parallelization. Incorporating these techniques would not only improve computational efficiency but also potentially eliminate the time-scale discrepancy between Isaac Sim and the rod engine.

Future work will therefore focus on upgrading \textit{DeformX} to adopt a stable Cosserat rod formulation, aiming to fundamentally resolve time-scale discrepancy issue while leveraging GPU acceleration for improved performance and scalability.

\section{Acknowledgment}
Generative AI tools were used for language refinement and code assistance. All content was reviewed, validated, and integrated by the authors. The authors take full responsibility for the originality and correctness of this work.

\bibliography{main}
\bibliographystyle{IEEEtran}

\clearpage
\newcommand{\rootloaded}{}
% =====================================================================
%  appendix.tex - Supplementary material for DeformX (IROS 2026, #3288)
% ---------------------------------------------------------------------
%  * \input{appendix} into main.tex; reuses its packages (graphicx,
%    booktabs, multirow, amsmath, xcolor) and the \todo macro.
%  * Place after \section{Conclusion and Discussion}, before
%    \section{Acknowledgment}. Standard floats; ieeeconf-compatible.
%  * Figures load from img/ and are column-width. distrubution.jpg is a
%    draft; jaccard_j_by_scene.png and the hit-apple montage are dense,
%    so consider a compact redesign (or a spanning figure*) if they read
%    too small at column width.
% =====================================================================

\section{Appendix}
\label{sec:appendix}

This appendix details the \textit{WireSeg-36k} dataset, the per-scene behaviour of the SAM3 baseline, RL formulation of the planar hit-target rope-swinging task, and a qualitative dynamic-manipulation demonstration (Hit Apple Task).

\subsection{Dataset Composition}
\label{app:composition}

\textit{WireSeg-36k} spans three scenario categories --- wire-on-plane, flying wires, and data center --- generated from $720$ independent simulation runs. The first two categories are further split by wire count ($\{2,4,8\}$) to vary clutter and inter-wire occlusion; Table~\ref{tab:dataset_status} summarises the per-scene composition.

\begin{figure}[h!]
\centering
\includegraphics[width=\linewidth]{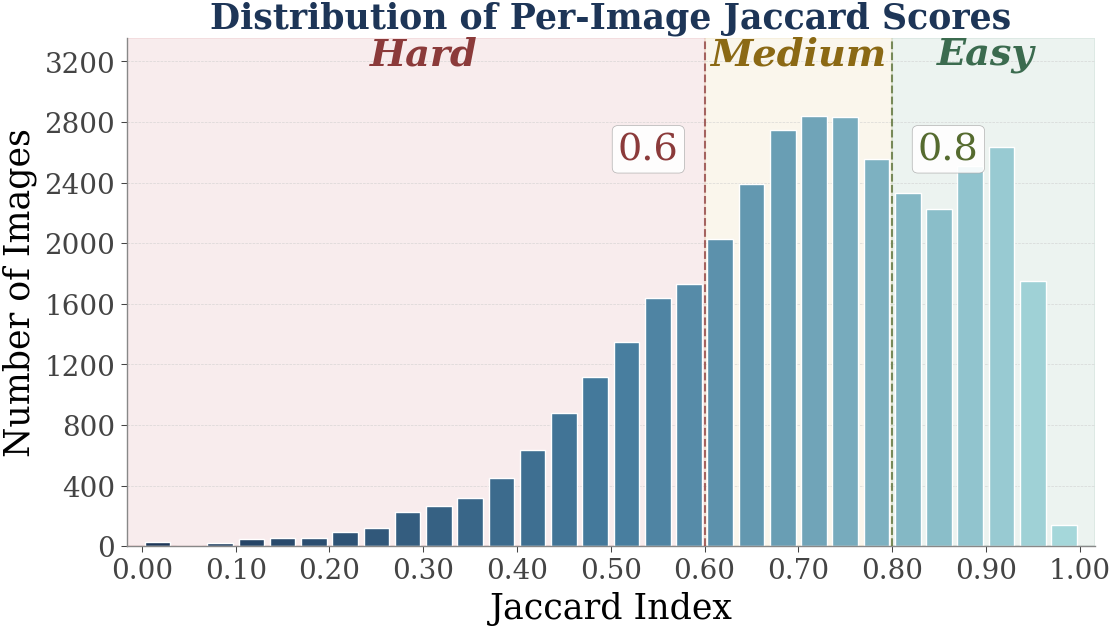}
\caption{All-data distribution of the per-image Jaccard score with the hard/medium/easy bands (thresholds $0.6$ and $0.8$).}
\label{fig:diff_dist}
\end{figure}

\begin{figure}[h!]
\centering
\includegraphics[width=\linewidth]{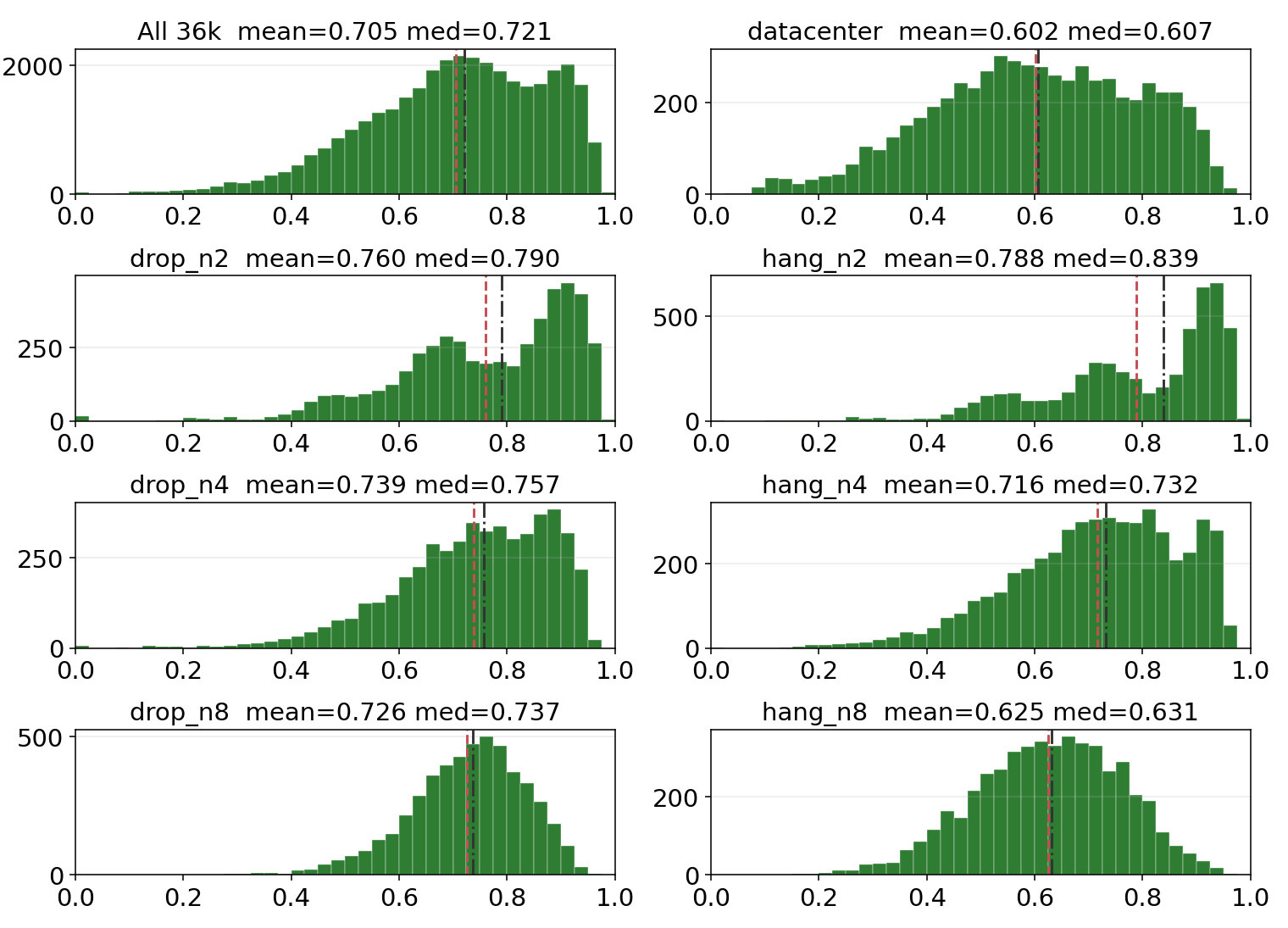}
\caption{Distribution of the per-image Jaccard score $J$ over the published \textit{WireSeg-36k} ($36{,}000$ images), using the off-the-shelf SAM3 baseline with the \texttt{``cable''} prompt. Thresholding $J$ at $0.6$ and $0.8$ stratifies the dataset into \textbf{hard} ($J<0.6$): $9{,}016$ images ($25.0\%$), \textbf{medium} ($0.6\le J<0.8$): $15{,}393$ ($42.8\%$), and \textbf{easy} ($J\ge0.8$): $11{,}591$ ($32.2\%$).}
\label{fig:jaccard_scene}
\end{figure}

\begin{table}[h!]
\centering
\scriptsize
\setlength{\tabcolsep}{3.5pt}
\renewcommand{\arraystretch}{1.2}
\begin{tabular}{llccccc}
\toprule
\textbf{Category} & \textbf{Scene} & \textbf{\#W} & \textbf{\#Img} & \textbf{Mean $J$} & \textbf{F1@75} &
\textbf{mAP@75} \\
\midrule
\multirow{3}{*}{Wire-on-plane} & \texttt{drop\_n2}   & 2    & 5{,}000 & 0.760 & 0.621 & 0.584 \\
                             & \texttt{drop\_n4}   & 4    & 5{,}000 & 0.739 & 0.609 & 0.550 \\
                             & \texttt{drop\_n8}   & 8    & 5{,}000 & 0.726 & 0.594 & 0.546 \\
\midrule
\multirow{3}{*}{Flying wires}  & \texttt{hang\_n2}   & 2    & 5{,}000 & 0.788 & 0.612 & 0.621 \\
                             & \texttt{hang\_n4}   & 4    & 5{,}000 & 0.716 & 0.508 & 0.472 \\
                             & \texttt{hang\_n8}   & 8    & 5{,}000 & 0.625 & 0.390 & 0.320 \\
\midrule
Data center                    & \texttt{datacenter} & 4,8,16 & 6{,}000 & 0.602 & 0.368 & 0.340 \\
\midrule
\textbf{All data}              &                     &      & \textbf{36{,}000} & 0.705 & 0.524 & 0.486 \\
\bottomrule
\end{tabular}
\caption{Status of \textit{WireSeg-36k}: per-scene composition and per-image Jaccard $J$, F1@75, and mAP@75 of the off-the-shelf SAM3 baseline (prompt \texttt{``cable''}); \#W is the number of wire in that scene simulation.}
\label{tab:dataset_status}
\end{table}

\subsection{Difficulty and Per-Scene Segmentation Quality}
\label{app:perscene}

% We grade per-image difficulty with the SAM3 baseline (text prompt \texttt{``cable''}) using the Jaccard index $J=\lvert\hat{M}\cap M\rvert/\lvert\hat{M}\cup M\rvert$ between the predicted and ground-truth wire masks. 

We grade per-image difficulty with the off-the-shelf SAM3 baseline (text prompt \texttt{``cable''}) using the Jaccard index (intersection-over-union) between predicted and ground-truth wire masks. Let $M$ be a ground-truth instance mask and $\hat{M}$ a predicted instance mask, each represented as the set of image pixels belonging to that instance. Their Jaccard index is
\begin{equation}
J(\hat{M}, M) \;=\; \frac{\lvert \hat{M}\cap M \rvert}{\lvert \hat{M}\cup M \rvert} \;\in\; [0,1],
\label{eq:jaccard}
\end{equation}
where $\cap$ and $\cup$ are the per-pixel intersection and union and $\lvert\cdot\rvert$ counts pixels, so $J=1$ denotes a perfect overlap and $J=0$ disjoint masks.

Fig.~\ref{fig:diff_dist} shows the all-data distribution of $J$ with the easy/medium/hard bands, and Fig.~\ref{fig:jaccard_scene} with Table~\ref{tab:dataset_status} break it down by scene. Two effects dominate: segmentation quality falls monotonically with wire count in both categories, and flying wires degrade faster than wire-on-plane at high density, as freely hanging cables overlap more in depth. The data-center scenes are hardest overall, limited by dense rack cabling and complex backgrounds rather than by simulation fidelity.

\begin{table}[h!]
\centering
\small
\setlength{\tabcolsep}{2pt}
\renewcommand{\arraystretch}{1.2}
\begin{tabular}{lccccc}
\toprule
\textbf{Prompt} & \textbf{AP} & \textbf{F1@75} & \textbf{$J$} & \textbf{zero-$J$} & \textbf{pred/GT} \\
\midrule
\textbf{cable}  & \textbf{0.352} & \textbf{0.524} & \textbf{0.705} & 25   & 1.94 \\
wire            & 0.328 & 0.508 & 0.694 & 61   & 1.89 \\
cord            & 0.310 & 0.478 & 0.683 & 141  & 1.79 \\
colorful cable  & 0.309 & 0.466 & 0.682 & 127  & 1.47 \\
foreground *     & 0.307 & 0.478 & 0.691 & \textbf{3} & 1.32 \\
colorful cords  & 0.303 & 0.453 & 0.693 & 248  & 1.41 \\
long cylinder   & 0.271 & 0.439 & 0.680 & 3441 & \textbf{1.08} \\
% per-scene       & 0.231 & 0.376 & 0.580 & 2289 & 1.14 \\
\bottomrule
\end{tabular}
\caption{Prompt sweep for the off-the-shelf SAM3 baseline over \textit{WireSeg-36k} ($36{,}000$ images); only the prompt is varied. AP is COCO mAP over IoU $0.50$--$0.95$, pooled dataset-wide. F1@75 (per-image F1 at IoU~$0.75$) and $J$ (per-image mean Jaccard) are computed per image and averaged. zero-$J$ counts images with $J = 0$ (no prediction or predicted masks but match no true wires); pred/GT is total number of predictions over total GT instances ($1.0$ ideal, $>1$ over-detection). Higher is better except zero-$J$ and pred/GT; Best per column in \textbf{bold}; * the foreground row is short for the prompt \texttt{``cables in the foreground''}.}
% \todo{The \texttt{per-scene} row falls below every fixed prompt on the per-image metrics (F1@75~$0.376$); a best-per-scene oracle should dominate them (notes cite F1@75~$0.529$). Confirm what it represents, or drop it.}}
\label{tab:prompt_sweep}
\end{table}

\begin{figure*}[t]
    \centering
    \includegraphics[width=\textwidth]{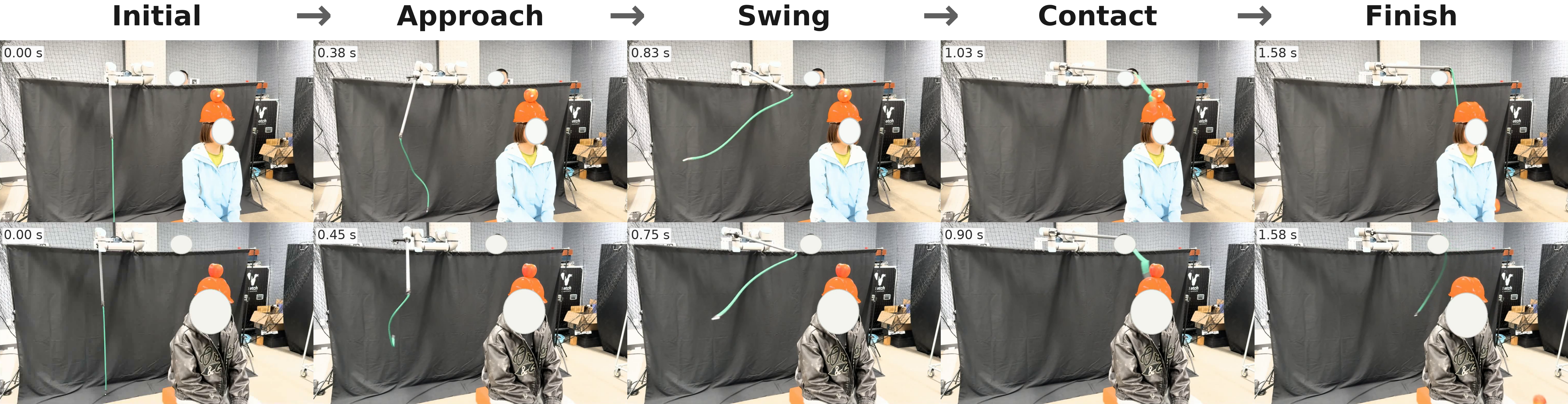}
    \caption{Hit-apple demonstration: temporal montage of two trials (rows)
    as a UR5e swings a rope through
    Initial$\,\to\,$Approach$\,\to\,$Swing$\,\to\,$Contact$\,\to\,$Finish
    to knock an apple off a person's head, actuating only joints 2, 3, and 4.}
    \label{fig:apple}
\end{figure*}

\subsection{Prompt Sensitivity of the SAM3 Baseline}
\label{app:prompt}

The main paper queries SAM3 with the single-token prompt \texttt{``cable''}; Table~\ref{tab:prompt_sweep} quantifies how much this choice matters. Using the same SAM3 off-the-shelf model (no fine-tuning), we sweep seven prompts and evaluate over all $36{,}000$ images. \texttt{``cable''} is the clear winner on almost every quality metric, with the more descriptive prompts trailing behind. The diagnostics expose a precision--recall trade-off: \texttt{``cable''} fires most aggressively --- pred/GT~$1.94$, roughly twice as many masks as the ground truth --- yet almost never fails outright ($25$ zero-$J$ images), whereas geometric or qualified prompts suppress over-detection at a steep recall cost. \texttt{``long cylinder''} nearly calibrates the instance count (pred/GT~$1.08$) but predicts no overlapping or no mask on $3{,}441$ images, and even its lenient AP$_{50}$ collapses to $0.379$; \texttt{``cable in the foreground''} fails on only three images yet still trails \texttt{``cable''} on AP. A fixed \texttt{``cable''} prompt is therefore a robust default, and prompt phrasing is not the main accuracy bottleneck.

\subsection{Reinforcement Learning Formulation}
\label{app:rl}

We use PPO to demonstrate that DeformX can support policy learning for
dynamic DLO manipulation. The task is inspired by the hit-target benchmark
of Chi \emph{et al.}~\cite{irp}, while the policy formulation and training
procedure are implemented independently.

\paragraph{Policy inputs and outputs.}
For the hit-target rope-swinging task, we actuate UR5e joints 1, 2, and 3
and denote the active-joint set by \(\mathcal A=\{1,2,3\}\). At each control
step, the policy observes the positions and velocities of the active robot
joints, the rope-tip position and velocity, the end-effector position, and
the target position:
\begin{equation}
\mathbf{s}_t =
\left[
\mathbf{q}_t^{\mathcal A},
\dot{\mathbf{q}}_t^{\mathcal A},
\mathbf{p}_t^{\mathrm{tip}},
\dot{\mathbf{p}}_t^{\mathrm{tip}},
\mathbf{p}_t^{\mathrm{ee}},
\mathbf{p}^{\mathrm{tar}}
\right].
\label{eq:rl_observation}
\end{equation}
This gives an 18-dimensional observation. The policy outputs three normalized
actions, which are scaled according to the joint-velocity limits and
accumulated as incremental joint-position commands. The resulting targets are
clipped to the valid joint ranges. During training, the policy operates in
closed loop and receives an updated robot and rope state at every control
step.

\paragraph{Reward and termination.}
Let \(d_t\) be the rope-tip distance to the target in the motion plane. We
organize the reward into five components:
\begin{equation}
r_t =
r_t^{\mathrm{near}}
+
r_t^{\mathrm{progress}}
+
r_t^{\mathrm{swing}}
+
r_t^{\mathrm{bonus}}
-
r_t^{\mathrm{regularization}}.
\label{eq:rl_reward}
\end{equation}
The proximity term rewards a small tip-to-target distance, while the progress
term rewards reductions in both the current distance and the closest distance
reached during the episode. The swing term encourages the rope tip to move
toward the target and promotes motion of the primary swing-driving joint.
Additional bonuses are given when the tip crosses a set of distance thresholds
and when the target is reached. We regularize the policy using penalties on
action magnitude, abrupt action changes, and episode duration. The reward is
clipped to avoid unusually large updates. An episode terminates when the
target is reached, the time limit is exceeded, or the rope tip leaves the
valid workspace.

\paragraph{Real-world execution.}
The trained policy is first used to generate successful joint-command
trajectories in simulation. Because the current physical setup does not
provide online estimates of the rope-tip state, these trajectories are
replayed open loop on the UR5e. The real-world experiment therefore evaluates
whether dynamic motions discovered in DeformX transfer to the physical
system, rather than evaluating closed-loop visual feedback control.
% With an online rope-tracking system, the same state-feedback policy could instead be executed closed loop on the robot.

\subsection{Dynamic Striking Demonstration}
\label{app:apple}

Beyond the quantitative hit-target benchmark, we include a qualitative demonstration of dynamic rope manipulation: a UR5e swings a rope to knock an apple off a person's head (Fig.~\ref{fig:apple}). The policy actuates only the second, third, and fifth joints of the arm, whipping the rope's free end onto the target with a horizontal velocity, and shows that the pipeline used for the hit-target task extends to dynamic striking not on the vertical plane.

Both tasks shape a reward on the rope's free end. The main-paper hit-target task rewards driving the tip toward the goal, with the return dominated by the minimum tip-to-goal distance $d_{\min}$ of Eq.~\ref{eq:dmin}. The striking demonstration instead rewards reaching the target with large horizontal tip velocity, encouraging a forceful, whip-like impact rather than a gentle touch.

\end{document}